\documentclass[11pt,a4paper]{article}
\usepackage[hyperref]{acl2020}
\usepackage{times}
\usepackage{latexsym}

\usepackage{graphicx}

\usepackage{microtype}

\usepackage{helvet}
\usepackage{courier}
\usepackage{url}
\usepackage{mathptmx}
\usepackage[hyphenbreaks]{breakurl}
\usepackage{xcolor}
\usepackage{bibentry}
\usepackage{stfloats}
\usepackage{multirow}
\usepackage{amssymb}
\usepackage{pifont}
\usepackage{amsmath}
\allowdisplaybreaks
\usepackage{array}
\DeclareMathAlphabet{\mathcal}{OMS}{cmsy}{m}{n}
\usepackage{enumitem}
\usepackage{stmaryrd}

\usepackage{booktabs}
\usepackage{subcaption}

\usepackage{algorithm}
\usepackage{algorithmic}
\renewcommand{\algorithmiccomment}[1]{\bgroup\hfill~#1\egroup}
\usepackage{footnote}

\aclfinalcopy % Uncomment this line for the final submission

\title{ScaleVLAD: Improving Multimodal Sentiment Analysis via \\Multi-Scale Fusion of Locally Descriptors}
\author{
	Huaishao Luo\textsuperscript{\rm 1}\thanks{~~This work was done during the first author's internship in MSR Asia}~,
	Lei Ji\textsuperscript{2},
	Yanyong Huang\textsuperscript{\rm 3},
	Bin Wang\textsuperscript{\rm 4},
	Shenggong Ji\textsuperscript{\rm 5},
	Tianrui Li\textsuperscript{1} \\
	\textsuperscript{\rm 1}Southwest Jiaotong University, Chengdu, China\\
	{\tt huaishaoluo@gmail.com, trli@swjtu.edu.cn} \\
	\textsuperscript{\rm 2}Microsoft Research Asia, Beijing, China \\
	\textsuperscript{\rm 3}Southwestern University of  Finance and Economics, Chengdu, China \\
	\textsuperscript{\rm 4}Ocean University of China, Qingdao, China \\
	\textsuperscript{\rm 5}Tencent, Shenzhen, China \\
	{\tt leiji@microsoft.com, huangyy@swufe.edu.cn} \\
	{\tt binwang9545@ouc.edu.cn }
	{\tt shenggongji@163.com }
} 
	
\date{}

\begin{document}
	\maketitle
	\begin{abstract}
		Fusion technique is a key research topic in multimodal sentiment analysis. The recent attention-based fusion demonstrates advances over simple operation-based fusion. However, these fusion works adopt single-scale, i.e., token-level or utterance-level, unimodal representation. Such single-scale fusion is suboptimal because that different modality should be aligned with different granularities. This paper proposes a fusion model named \textbf{ScaleVLAD} to gather multi-\textbf{Scale} representation from text, video, and audio with shared \textbf{V}ectors of \textbf{L}ocally \textbf{A}ggregated \textbf{D}escriptors to improve unaligned multimodal sentiment analysis. These shared vectors can be regarded as shared topics to align different modalities. In addition, we propose a self-supervised shifted clustering loss to keep the fused feature differentiation among samples. The backbones are three Transformer encoders corresponding to three modalities, and the aggregated features generated from the fusion module are feed to a Transformer plus a full connection to finish task predictions. Experiments on three popular sentiment analysis benchmarks, IEMOCAP, MOSI, and MOSEI, demonstrate significant gains over baselines.
	\end{abstract}  
	
	\section{Introduction}
	Multimodal Sentiment Analysis (MSA) has been a hot research direction with the increasing number of user-generated videos available on online platforms such as YouTube and Facebook in recent years \cite{poria2020beneath, Tsai2019Multimodal, Zadeh2017Tensor}.
	\begin{figure}[!th]
		\centering
		\begin{subfigure}[t]{0.98\linewidth}
			\centering
			\includegraphics[width=1.0\linewidth]{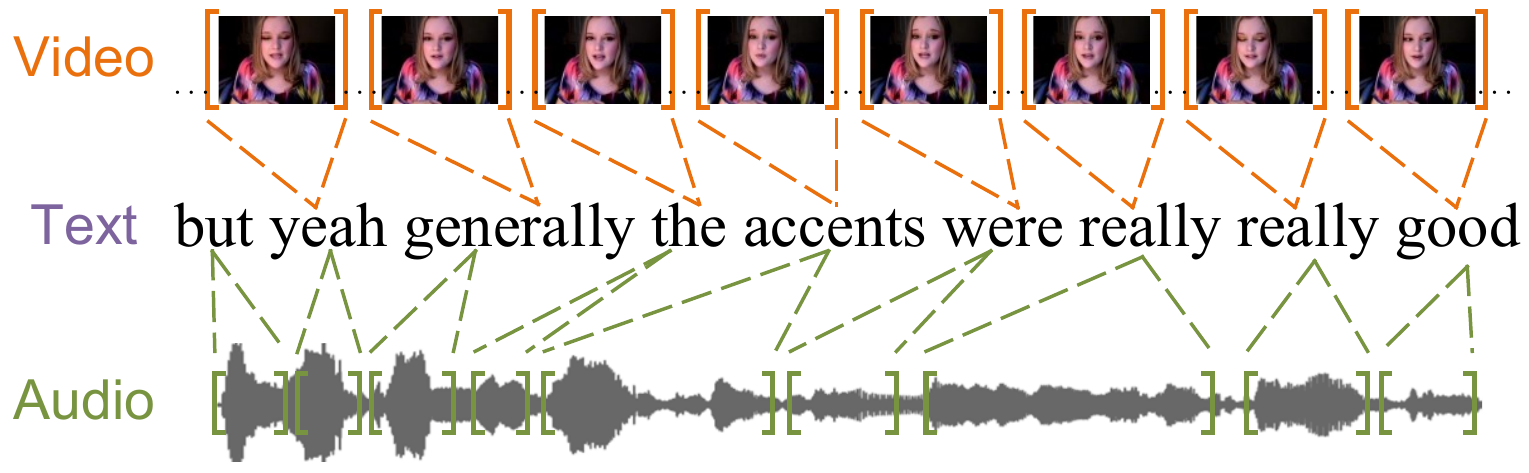}
			\caption{Single-scale alignment}
			\label{fig:singlescale_alignment}
		\end{subfigure} \\
		\begin{subfigure}[t]{0.98\linewidth}
			\centering
			\includegraphics[width=1.0\linewidth]{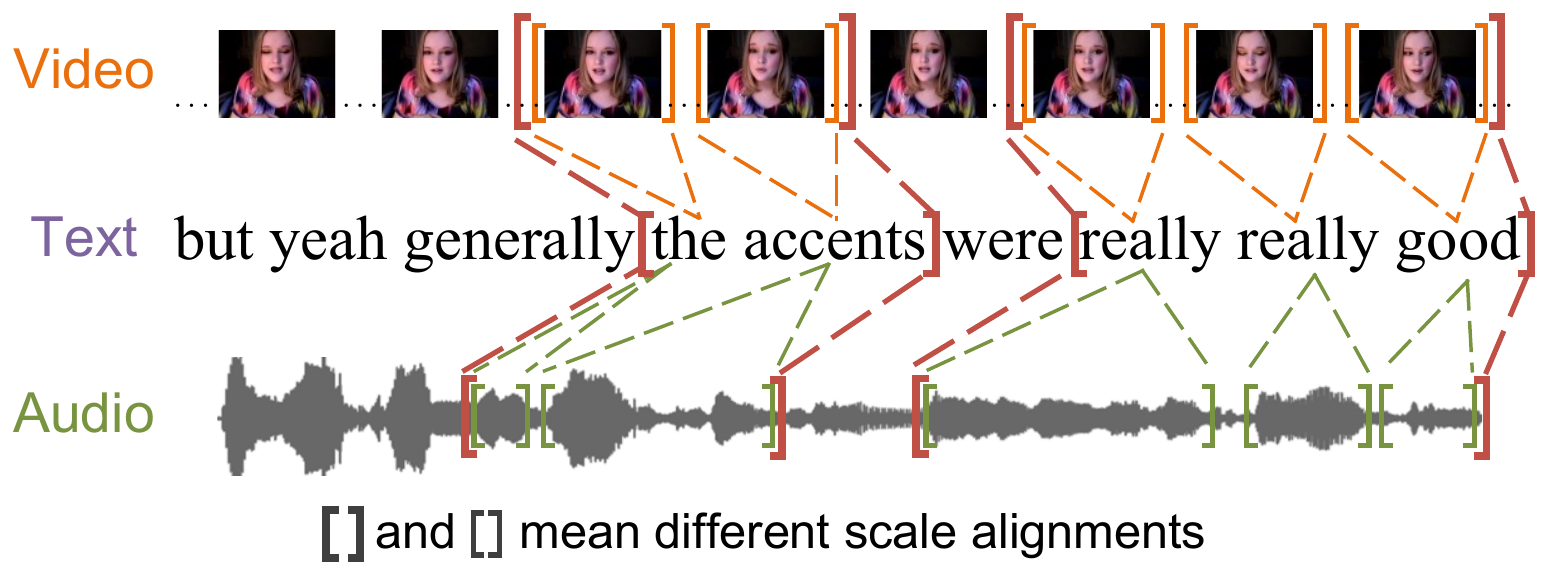}
			\caption{Multi-scale alignment}
			\label{fig:scale_alignment}
		\end{subfigure}
		\caption{Illustration of alignment between text, video, and audio. (\ref{sub@fig:singlescale_alignment}) single-scale alignment. (\ref{sub@fig:scale_alignment}) our multi-scale alignment.}
		\label{fig:alignment} 
	\end{figure}
	Its main objective is to identify sentiment and emotion with multimodal signals such as textual, visual, and acoustic information. Compared with unimodal sentiment analysis, multimodal fusion can provide more comprehensive information and capture more emotional characteristics, which leads to robust and salient improvements \cite{Yang2020CMBERT, Yu2021SelfMM}. For example, to judge the sentiment of \emph{this movie is sick} is a non-trivial task due to the existing language ambiguity only from this sentence, and if given the acoustic and visual modalities, e.g., a loud voice and a smile, this sentence will certainly be predicted as positive \cite{Zadeh2017Tensor, Wang2020TransModality}.

	There are two main components in multimodal sentiment analysis: unimodal representation and information fusion. For the unimodal representation, there are some off-the-shelf methods. These methods are elaborate and specialized for each modality or can be improved with pretraining on extra pure datasets, e.g., MFCC for audio and BERT encoding for text \cite{Devlin2019BERT}. Thus, multimodal information fusion is the key to affect performance \cite{poria2020beneath,Zhang2020Multimodal}. Most of the works focus on investigating effective multimodal fusion. These fusion methods can be categorized into including but not limited to simple operation-based \cite{Poria2016Convolutional}, attention-based \cite{zadeh2018multi, Gu2018Multimodal, Akhtar2019Multi, han2021bi, Rahman2020Integrating}, tensor-based \cite{Zadeh2017Tensor}, translation-based \cite{Pham2019Found, Wang2020TransModality, mai2020modality}, GANs-based \cite{peng2019cm}, graph-based \cite{Yang2021MTAG}, and routing-based methods \cite{Tsai2020Multimodal}. The fusion target is to learn a modality-invariant embedding space, then use the modality-invariant feature or integrate the modality-invariant with modality-specific features to finish the final prediction.
	
	However, most of the fusion methods either adopt the token-level or the utterance-level unimodal representation. Such a single-scale fusion is suboptimal because different modalities need to align with different granularities. For example, the `really really good' shown in Figure \ref{fig:alignment}. The single-scale alignment of the three tokens can not capture the intense emotion. Instead, they should be regarded as an entirety shown in Figure \ref{fig:scale_alignment}. Besides, the visual and acoustic features do not have apparent semantic boundaries due to variable sampling rates, leading to inherent data non-alignment for each modality \cite{Tsai2019Multimodal}. Although the attention-based methods can make each token in one modality cover long-range contexts in other modalities, they are still single-scale alignment and can not capture \emph{many tokens-to-many tokens} relationship.
	
	To this end, we propose a multi-scale fusion method called \textbf{ScaleVLAD} to gather multi-\textbf{Scale} representation from text, video, and audio with shared \textbf{V}ectors of \textbf{L}ocally \textbf{A}ggregated \textbf{D}escriptors to address the \emph{unaligned} multimodal sentiment analysis. Instead of detecting the boundary of different semantic scales in each modality, ScaleVLAD utilizes learnable shared latent semantic vectors to select and aggregate the modality features automatically. These latent semantic vectors, regarded as different semantic topics, are shared across different modalities and scales. Thus they can reduce the semantic gap between modalities and align various scale features naturally. In our implementation, we use three Transformer-based modules \cite{vaswani2017attention} to extract unimodal representation from text, video, and audio, respectively. Then, the unimodal feature sequences are fed to the ScaleVLAD module with different scales of shifted windows. The aggregated features from the ScaleVLAD module are used to predict the final output via a Transformer and a full connection layer. Figure \ref{fig:main_structure} shows the main structure of the proposed ScaleVLAD. Besides, to keep the differentiation of the fused feature among samples and leverage label information effectively, we propose a self-supervised shifted clustering loss to train the model jointly. This loss will pull clusters of samples belonging to the same category (or close score) together in embedding space. The contribution of this paper can be summarized as follows:

	1) We propose a multi-scale fusion method ScaleVLAD to address the unaligned multimodal sentiment analysis. It is a flexible approach to fuse unimodal representation with a multi-scale perspective.
	
	2) We propose a self-supervised shifted clustering loss to keep the fused feature differentiation among samples and leverage label information effectively.
	
	3) We report new records on three benchmark datasets, including IEMOCAP \cite{Busso2008IEMOCAP}, CMU-MOSI \cite{zadeh2016multimodal}, and CMU-MOSEI \cite{Zadeh2018MOSEI}. Extensive experiments validate the effectiveness of ScaleVLAD.

	\section{Related Works} 
	\subsection{Multimodal Sentiment Analysis}
	In recent years, multimodal sentiment analysis has become a popular research topic as the increasing of user-generated multimedia data on online communities, blogs, and multimedia platforms. It mainly focuses on integrating multiple heterogeneous resources, such as textual, visual, and acoustic signals to comprehend varied human emotions \cite{Morency2011Towards, poria2020beneath}. Previous researchers mainly focus on unimodal representation learning and multimodal fusion. For the unimodal representation, \cite{Hazarika2020MISA} attempted to factorize modality features in joint spaces and presented modality-invariant and modality-specific representations across different modalities. \cite{Yu2021SelfMM} designed a unimodal label generation strategy based on the self-supervised approach to acquire information-rich unimodal representations by learning one multimodal task and three unimodal subtasks. \cite{Wang2019Words} constructed a recurrent attended variation embedding network to model the fine-grained structure of nonverbal sub-word sequences and dynamically shift word representations based on nonverbal cues.

	For the multimodal fusion, the previous methods can be divide into simple operation-based \cite{Poria2016Convolutional}, attention-based \cite{zadeh2018multi, Gu2018Multimodal, Akhtar2019Multi, han2021bi, Rahman2020Integrating}, tensor-based \cite{Zadeh2017Tensor,Verma2019DeepCU,Verma2020Deep}, translation-based \cite{Pham2019Found, Wang2020TransModality, mai2020modality}, GANs-based \cite{peng2019cm}, graph-based \cite{Yang2021MTAG,mai2020analyzing}, and routing-based methods \cite{Tsai2020Multimodal}, etc. Some works assumed the given multimodal sequences are aligned with each word's boundary \cite{Pham2019Found, Gu2018Multimodal, Dumpala2019Audio, Rahman2020Integrating}. However, some modalities, e.g., video and audio, exist inherent data non-alignment due to variable sampling rates. Thus, modeling unaligned multimodal sequences is more flexible and practical. \cite{Tsai2019Multimodal, Yang2020CMBERT, Siriwardhana2020Jointly} used multiple cross-modal Transformers to model unaligned multimodal language sequences. \cite{Yang2021MTAG} proposed a parameter-efficient and interpretable graph-based neural model by integrating an efficient trimodal-temporal graph fusion operation and dynamic pruning technique.
	\begin{figure*}[tp]
		\centering
		\includegraphics[width=0.98\textwidth]{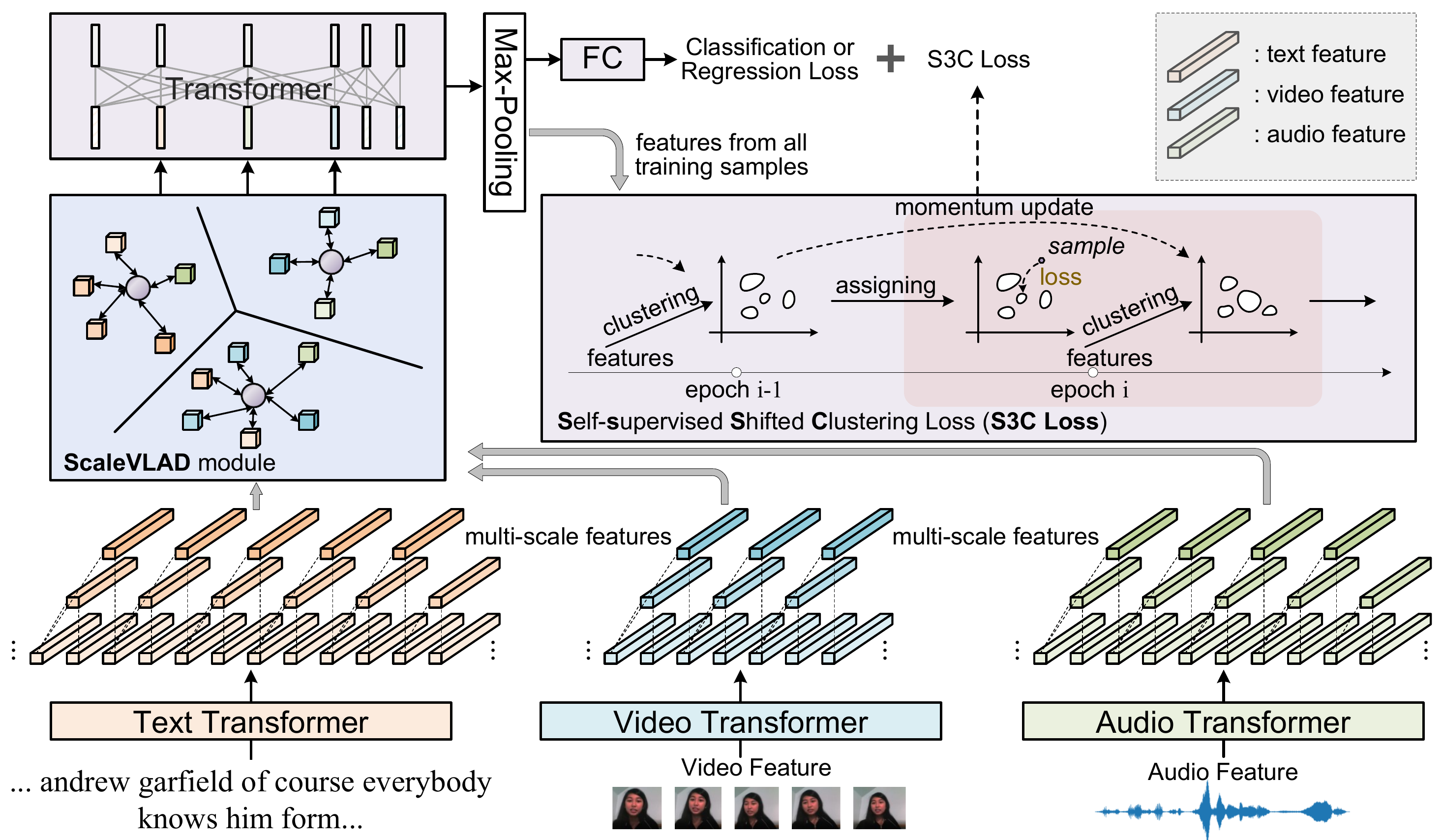} 
		\caption{The main structure of our ScaleVLAD, which comprises four components, including three unimodal encoders, and a fusion module. The model is trained with a task-related loss and an extra clustering loss.} 
		\label{fig:main_structure} 
	\end{figure*}

	This paper aims at unaligned multimodal sentiment analysis. Unlike previous studies adopting the token-level or the utterance-level unimodal representation, we propose a multi-scale fusion method to align different granularity information from multiple modalities.

	\subsection{VLAD, Vector of Locally Aggregated Descriptors} 
	The Vector of Locally Aggregated Descriptors (VLAD) \cite{Jegou2010Aggregating, Arandjelovic2013All} has achieved great impacts in aggregating discriminative features for various scenarios, including video retrieval and video classification. NetVLAD \cite{Arandjelovic2016NetVLAD} extending from the VLAD is an end-to-end differentiable layer that could be readily plugged into many existing neural models. This paper borrows the idea of VLAD and NetVLAD to align different modalities, e.g., text, video, and audio, instead of using to be as a discriminative feature learner. \cite{Wang2021CVPR} has a similar motivation that leverages NetVLAD to reduce the gap of locally learned features from texts and videos. However, their objective is for text-video local similarity matching, and we have a different target. Besides, we introduce multi-scale features for enhanced fusion performance. \cite{hausler2021patchnetvlad} also presents a multi-scale fusion by deriving patch-level features from NetVLAD residuals. However, it is designed for place recognition and only on visual modality. We focus on unaligned multimodal sentiment analysis and involves text, video, and audio modalities.

	\section{Framework}
	\label{section:framework}
	Given a set of multimodal signals including text $\mathcal{T}$, video clips $\mathcal{V}$, and audios $\mathcal{A}$, the target is to predict their sentiment. Specifically, these signals can be regarded as a set of triplets $(T_i, V_i, A_i)$, where $T_i \in \mathcal{T}$, $V_i \in \mathcal{V}$ and $A_i \in \mathcal{A}$. The $T_i$, $V_i$, and $A_i$ contain a sequence of tokens, respectively, such that $T_i = \big\{t_i^{j}|j\in[1, |T_i|]\big\}$, $V_i = \big\{\boldsymbol{v}_i^{j}|j\in[1,|V_i|]\big\}$, and $A_i = \big\{\boldsymbol{a}_i^{j}|j\in[1,|A_i|]\big\}$, where $t_i^{j}$ is word token, $\boldsymbol{v}_i^{j}$ is visual feature, and $\boldsymbol{a}_i^{j}$ denotes acoustic feature. We regard the visual features and acoustic features as tokens for a consistent description with the word tokens. Multimodal sentiment analysis aims to learn a function $f(T_i, V_i, A_i)$ to get the sentiment score or emotion category. The function learning can be regarded as either a regression or a classification task. 
	
	Figure \ref{fig:main_structure} demonstrates our framework. We focus on the multi-scale fusion module and a training loss, S3C loss, in this paper. Besides, three unimodal encoders, a text encoder, a video encoder, and an audio encoder, are also introduced in detail in this section.

	\subsection{Modality Representation Learning}
	The unimodality representation is the footstone of this model and will affect the performance of the subsequential fusion module. We use Transformer \cite{vaswani2017attention} with different layers to encode original text $T_i$, raw video feature sequence $V_i \in \mathbb{R}^{|V_i| \times \hat{d}_{v}}$, and raw audio feature sequence $A_i \in \mathbb{R}^{|A_i| \times \hat{d}_{a}}$, where $\hat{d}_{v}$ and $\hat{d}_{a}$ are the dimensions of the raw feature. The raw video feature and raw audio feature are extracted with pretrained toolkits following previous works \cite{Zadeh2017Tensor, Yu2021SelfMM}. For the text encoder, we use the pretrained 12-layers BERT \cite{Devlin2019BERT} and 12-layers T5 \cite{Raffel2020t5} to extract text representation $\mathcal{F}_{T_i} \in \mathbb{R}^{|T_i| \times d_{t}}$ since the tremendous success of the pre-trained language model on many downstream NLP tasks, where $d_{t} = 768$ is the dimension of the text representation.
	\begin{align}
		\mathcal{F}_{T_i} = \texttt{Transformer}_\texttt{T}(T_i),
	\end{align}
	where $\texttt{Transformer}_\texttt{T}$ means the Transformer-based text encoder, e.g., BERT and T5 in our implementation.

	Similarly, the video feature sequence $\mathcal{F}_{V_i} \in \mathbb{R}^{|V_i| \times d_{v}}$ and audio feature sequence $\mathcal{F}_{A_i} \in \mathbb{R}^{|A_i| \times d_{a}}$ can be calculated with $V_i$ and $A_i$ respectively as follows,
	\begin{align}
		\mathcal{F}_{V_i} = \texttt{Transformer}_\texttt{V}(V_i), \\
		\mathcal{F}_{A_i} = \texttt{Transformer}_\texttt{A}(A_i),
	\end{align}
	where $\texttt{Transformer}_\texttt{V}$ and $\texttt{Transformer}_\texttt{A}$ are Transformer-based video encoder and Transformer-based audio encoder, respectively, both of them are randomly initialized. $d_{v}$ and $d_{a}$ are the dimension of the video feature and audio feature, respectively.

	\subsection{ScaleVLAD Module}
	\label{sec_scalevland_module}
	After generating the unimodality representation, the framework comes to the fusion module. We propose a multi-scale fusion method to cover different granularities of unimodality representation in this paper. Different full connection layers are used for the generated $\mathcal{F}_{T_i}$, $\mathcal{F}_{V_i}$, and $\mathcal{F}_{A_i}$ to map the hidden size to a common size $d_s$ before the following modules if their current hidden sizes are not equal to this value. When considering the fusion of the three unimodality features, especially with different granularities, a core problem is aligning different semantic units. However, the semantic unit of each unimodality has no clear alignment boundary and can not be fused directly. A feasible approach is to assume some shared semantic vectors among these unimodality features and align them to these shared anchors. Such shared vectors can be regarded as shared topics and can also be shared across different unimodality scales. 
	
	Motivated by this spirit and Inspired by the VLAD and NetVLAD, we propose a ScaleVLAD module to fuse different unimodality representations. The different scale information of unimodality is generated by \texttt{mean pooling} with different kernel size (the stride size is the same as the kernel size) in our implementation. Specifically, for $m$-scale unimodality representation $\mathcal{F}_{M_i}, M \in \{T, V, A\}$, the scaled features can be denoted as $\mathcal{F}_{M_i}^{(m)} = \{\boldsymbol{f}_j^{(m)} | j \in [1, |\mathcal{F}_{M_i}^{(m)}|]\}$, where $\boldsymbol{f}_j^{(m)}$ is generated via \texttt{mean pooling} with kernel size $m$. The $\mathcal{F}_{M_i}^{(m)}$ is equal to $\mathcal{F}_{M_i}$ when $m=1$. Assuming there are $K$ shared semantic vectors $\{\boldsymbol{c}_k | k \in [1, K]\}$ with $d_s$ dimension. The similarity between the $m$-scale feature $\boldsymbol{f}_j^{(m)}$ and the shared vectors can be calculated by dot-product operation following \cite{Arandjelovic2016NetVLAD},
	\begin{align}
		w_{ij}^{(m)} = \frac{\exp (\boldsymbol{f}_i^{(m)}\boldsymbol{c}_j^\top + b_j)}{ \sum_{k=1}^{K} \exp (\boldsymbol{f}_i^{(m)}\boldsymbol{c}_k^\top + b_k )}, \label{eq_wij}
	\end{align}
	where $b_j$ and $b_k$ are learnable biases, the shared semantic vectors are jointly learned with the whole model. Then the aggregated feature on each vector can be generated as follows,
	\begin{align}
		\hat{\boldsymbol{r}}_j^{(m)} = &\sum\nolimits_{i=1}^{|\mathcal{F}_{M_i}^{(m)}|} w_{ij}^{(m)} (\boldsymbol{f}_i^{(m)} - \hat{\boldsymbol{c}}_j), \\
		\boldsymbol{r}_j^{(m)} = & \hat{\boldsymbol{r}}_j^{(m)} / {\lVert \hat{\boldsymbol{r}}_j^{(m)} \rVert _2},
	\end{align}
	where $\hat{\boldsymbol{c}}_j$ has the same size as $\boldsymbol{c}_j$, and using two groups of similar vectors increases the adaptation capability as described in \cite{Arandjelovic2016NetVLAD}. The output $\boldsymbol{r}_j^{(m)}$ can be regarded as the aligned feature for unimodality with $m$-scale. Thus, the aggregated feature corresponding to $\mathcal{F}_{M_i}$ can be generated as follows,
	\begin{align}
		&\hat{\boldsymbol{u}} = \texttt{stack}([\boldsymbol{r}_1^{(m)}, \boldsymbol{r}_2^{(m)}, \cdots, \boldsymbol{r}_K^{(m)}]), \\
		&\boldsymbol{u}_{M_i}^{(m)} = \texttt{LN}(\texttt{GELU}(\hat{\boldsymbol{u}}\mathbf{W}\!_{M} + \boldsymbol{b}_M)),
	\end{align}
	where $\texttt{stack}$ is a stack operation and $\hat{\boldsymbol{u}} \in \mathbb{R}^{Kd_s}$, $\mathbf{W}\!_{M} \in \mathbb{R}^{Kd_s \times d_s}$ and $\boldsymbol{b}_M \in \mathbb{R}^{d_s}$ ($M \in \{T, V, A\}$) are learnable weights and biases, $\texttt{GELU}$ and $\texttt{LN}$ are GELU activate function \cite{hendrycks2016gaussian} and Layer Normalization operation \cite{Ba2016Layer}, respectively.
	
	The fusion and prediction are conducted on the multi-scale aggregated features $\boldsymbol{u}_{M_i}^{(m)}$. We stack all the representation with different scales $m_1, m_2, \dots$ together to get representation matrix, $R_i=[\boldsymbol{u}_{T_i}^{(m_1)}, \boldsymbol{u}_{V_i}^{(m_1)}, \boldsymbol{u}_{A_i}^{(m_1)}, \boldsymbol{u}_{T_i}^{(m_2)}, \dots, \bar{\boldsymbol{f}}_{T_i}, \bar{\boldsymbol{f}}_{V_i}, \bar{\boldsymbol{f}}_{A_i}] \in \mathbb{R}^{(3 \cdot |m|+3) \times d_s}$, where $|m|$ means the number of scales, $\bar{\boldsymbol{f}}_{M_i}$ $(M \in \{T, V, A\})$ is the $\texttt{mean pooling}$ result on $\mathcal{F}_{M_i}$. After obtaining $R_i$, a randomly initialized Transformer encoder $\texttt{Transformer}_\texttt{F}$ is utilized to interact the learned multi-scale representation:
	\begin{align}
		\hat{R}_i = \texttt{Transformer}_\texttt{F}(R_i). \label{fusion_transfomer}
	\end{align}
	Finally, the score or probability can be calculated as,
	\begin{align}
		&\hat{\boldsymbol{r}} = \texttt{max-pooling}(\hat{R}_i), \label{fusion_feature}\\
		&\boldsymbol{o}_i = \hat{\boldsymbol{r}}\mathbf{W}\!_{r} + \boldsymbol{b}_r, \label{output}
	\end{align}
	where $\hat{\boldsymbol{r}} \in \mathbb{R}^{d_s}$ is the \texttt{max pooling} result of $\hat{R}_i$, $\mathbf{W}_{r} \in \mathbb{R}^{d_s \times c}$ and $\boldsymbol{b}_r \in \mathbb{R}^{c}$ are learnable weights and biases, $c$ is the number of categories for classification task or 1 for regression task.

	\subsection{S3C Loss, Self-supervised Shifted Clustering Loss}
	Beyond proposing the ScaleVLAD module to capture and align different granularities of unimodality representation, we proposed an extra self-supervised shifted clustering loss (S3C Loss) to keep the differentiation of the fused feature among samples and to leverage label information effectively. 
	
	For the fusion feature $\hat{\boldsymbol{r}}$ of each sample from Eq. (\ref{fusion_feature}), we first perform $k$-means to obtain $C$ clusters\footnote{We use the Faiss (https://github.com/facebookresearch/faiss) to finish clustering in our implementation.}. We refer to the $i$-th cluster center as $\boldsymbol{z}_i \in \mathbb{R}^{d_s}$ and refer to all cluster centers as a matrix $Z \in \mathbb{R}^{C \times d_s}$. The clustering operation is calculated on all representations of training samples at each epoch beginning. For the same sample in the running epoch, we assign its cluster center index $i$ as a classified label. The S3C Loss can be obtained as follows,
	\begin{align}
		&\boldsymbol{p} = \texttt{softmax}(Z\hat{\boldsymbol{r}}), \\
		&\mathcal{L}_{s3c} = - \frac{1}{N} \sum_{i=1}^N \left(\mathbb{I}(i)(\log (\boldsymbol{p}))^{\top} \right),
	\end{align}
	where $\mathbb{I}(i)$ means the one-hot vector with length $C$ and its $i$-th value is 1, $N$ is the number of training samples.
	
	This loss is self-supervised but the clustering centers are not stable at the beginning of the training stage. So we set a start epoch $s_{s3c}$ to train with $\mathcal{L}_{s3c}$ instead of optimizing it from training beginning. Such a setting makes the features used for clustering semantically relate to the group-truth labels. To make the cluster centers stable, we adopt a shifted update with a momentum parameter $\alpha$ as $Z^{(t)} = \alpha Z^{(t-1)} + (1 - \alpha ) Z$ and use $Z^{(t)}$ to replace $Z$ at each iteration. The $\alpha$ is set as a constant of 0.99 in our experiments. The clustering loss makes the fusion features differentiate in the embedding space. 
	
	To improve the weak robustness caused by the unknown ground-truth cluster number of the fusion space, we design multiple clustering, e.g., with $C_1$ clusters and $C_2$ clusters. Thus, the $\mathcal{L}_{s3c}$ will be replaced by $\mathcal{L}_{s3c} = \mathcal{L}_{s3c}^{(C_1)} + \mathcal{L}_{s3c}^{(C_2)} + \dots$, where $\mathcal{L}_{s3c}^{(C_i)}$ means $\mathcal{L}_{s3c}$ with $C_i$ clusters.

	\subsection{Training Objectives}
	The overall objective of the model is to minimize:
	\begin{align}
		\mathcal{L} = \mathcal{L}_{task} + \mathcal{L}_{s3c},
	\end{align}
	where $\mathcal{L}_{s3c}$ is the S3C Loss, and $\mathcal{L}_{task}$ is the task loss. The task loss has different formulations for the classification task and regression task. For the classification task, we use cross-entropy error with $\boldsymbol{o}_i$ in Eq. (\ref{output}) as $\mathcal{L}_{task} = - \frac{1}{N} \sum_{i=1}^N (\mathbb{I}(y_i)(\log (\boldsymbol{o}_i))^\top)$, where $\mathbb{I}(y_i)$ means the one-hot vector of $y_i$. For the regression task, we use mean MSE as the training objective as $\mathcal{L}_{task} = \frac{1}{N} \sum_{i=1}^N (\lVert y_i - \boldsymbol{o}_i \rVert_2^2)$. $y_i$ is the category for classification or the score for regression, and $N$ is the number of training samples.

	\section{Experiments}
	We conduct experiments to evaluate the effectiveness of the proposed framework. The datasets, experimental settings, and results are introduced in this section.

	\subsection{Datasets} 
	We evaluate our framework on three benchmark datasets, IEMOCAP \cite{Busso2008IEMOCAP}, CMU-MOSI \cite{zadeh2016multimodal}, and CMU-MOSEI \cite{Zadeh2018MOSEI}. These datasets provide unaligned language, visual, and acoustic signals for multimodal sentiment analysis.
	
	\noindent
	\textbf{IEMOCAP} IEMOCAP \cite{Busso2008IEMOCAP} consists of 10,000 videos for human emotion analysis. We follow \cite{Wang2019Words} and select four emotions (happy, sad, angry, and neutral) for emotion recognition. The task of this dataset is a multilabel task (e.g., a person can be sad and angry simultaneously). The metric used on this dataset is the binary classification accuracy (Acc) and the F1 score of the predictions.
	
	\noindent
	\textbf{CMU-MOSI} Multimodal Opinion Sentiment and Emotion Intensity \cite{zadeh2016multimodal} is sentence-level sentiment analysis and emotion recognition in online videos. CMU-MOSI contains 2,199 opinion video clips, each annotated with real-valued sentiment intensity annotations in the range [-3, +3]. We evaluate the model performances using various metrics following prior works: binary accuracy (BA), F1 score, mean absolute error (MAE) of the score, and the correlation of the prediction with humans (Corr).
	
	\noindent
	\textbf{CMU-MOSEI} The CMU-MOSEI dataset \cite{Zadeh2018MOSEI} improves over MOSI with a higher number of utterances, greater variety in samples, speakers, and topics. The dataset contains 23,453 annotated video segments (utterances), from 5,000 videos, 1,000 distinct speakers and 250 different topics. The metrics are the same as the CMU-MOSI. 
	\begin{table*}[!tb]
		% \vspace{.57cm}
		\setlength{\tabcolsep}{2pt}
		\centering
		\scalebox{1.}{
		\begin{tabular}{lcccccccc}
			\toprule
			\multirow{2}{*}{Pretrained} & \multicolumn{2}{c}{IEMOCAP} & \multicolumn{2}{c}{CMU-MOSI} & \multicolumn{2}{c}{CMU-MOSEI} \\ 
										& $\overline{\text{Acc}}\uparrow$ & $\overline{\text{F1}}\uparrow$ & BA$\uparrow$ & F1$\uparrow$ & BA$\uparrow$ & F1$\uparrow$ \\ \midrule
			BERT-Base					& \textbf{82.9} & \textbf{82.6} & 85.0/86.9 & 84.9/86.9 & 82.9/86.1 & 83.3/86.1 \\ 
			T5-Base						& 82.6 & 82.4 & \textbf{87.2/89.3} & \textbf{87.3/89.3} & \textbf{84.5/86.4} & \textbf{84.7/86.3} \\ 
			\bottomrule
		\end{tabular}
		}
		\caption{\textbf{Text Encoder.} T5-Base has better performance than BERT-Base summarily. $\overline{\text{Acc}}$ and $\overline{\text{F1}}$ of IEMOCAP are the average values of Acc and F1, respectively.}
		\label{tab:result_of_pretrain}
	\end{table*}
	\begin{table*}[!tp]
		\setlength{\tabcolsep}{3pt}
		\centering
		\scalebox{1.}{
		\begin{tabular}{ccccccccc}
			\toprule
			\multirow{2}{*}{Scale} & \multicolumn{2}{c}{IEMOCAP} & \multicolumn{2}{c}{CMU-MOSI} & \multicolumn{2}{c}{CMU-MOSEI} \\ 
									& $\overline{\text{Acc}}\uparrow$ & $\overline{\text{F1}}\uparrow$ & BA$\uparrow$ & F1$\uparrow$ & BA$\uparrow$ & F1$\uparrow$ \\ \midrule
			1						& 82.2 & 82.1 & 86.4/88.8 & 86.2/88.8 & 83.3/85.9 & 83.3/85.9 \\ 
			1,2						& 82.5 & 82.2 & 86.3/88.9 & 86.3/88.8 & 83.2/86.2 & 83.6/86.2 \\ 
			1,3						& 81.9 & 81.8 & 86.4/88.9 & 86.4/88.9 & 83.5/86.3 & 83.6/86.2 \\ 
			1,2,3					& 82.0 & 81.9 & 86.7/89.0 & 86.6/89.0 & 84.3/\textbf{86.4} & 84.0/86.2 \\ 
			1,2,10					& \textbf{82.6} & \textbf{82.4} & 86.7/89.0 & 86.8/89.1 & 84.0/86.2 & 84.2/\textbf{86.3} \\ 
			1,2,3,10				& 82.1 & 81.9 & \textbf{87.2/89.3} & \textbf{87.3/89.3} & \textbf{84.5/86.4} & \textbf{84.7/86.3} \\ 
			\bottomrule
		\end{tabular}
		}
		\caption{\textbf{Multi-scale Fusion}. Fusing different scale features improve the performance.}
		\label{tab:result_of_scale}
	\end{table*}
	\begin{table*}[!tp]
		\setlength{\tabcolsep}{3pt}
		\centering
		\scalebox{1.}{
		\begin{tabular}{ccccccccc}
			\toprule
			\multirow{2}{*}{Cluster} & \multicolumn{2}{c}{IEMOCAP} & \multicolumn{2}{c}{CMU-MOSI} & \multicolumn{2}{c}{CMU-MOSEI} \\ 
									& $\overline{\text{Acc}}\uparrow$ & $\overline{\text{F1}}\uparrow$ & BA$\uparrow$ & F1$\uparrow$ & BA$\uparrow$ & F1$\uparrow$ \\ \midrule
			10						& 82.1 & 82.1 & 86.3/88.9 & 86.2/88.8 & 83.8/86.4 & 84.1/86.3 \\ 
			15						& 82.3 & 82.1 & 86.4/88.7 & 86.4/88.7 & 83.6/\textbf{86.5} & 83.9/86.4 \\ 
			20						& 82.5 & 82.2 & 86.3/88.6 & 86.2/88.6 & 84.1/\textbf{86.5} & 84.4/\textbf{86.5} \\ 
			10,15					& \textbf{82.6} & \textbf{82.4} & \textbf{87.2/89.3} & \textbf{87.3/89.3} & \textbf{84.5}/86.3 & \textbf{84.7}/86.2 \\ 
			15,20					& 82.3 & 82.1 & 85.4/87.9 & 85.3/87.9 & \textbf{84.5}/86.4 & \textbf{84.7}/86.3 \\ 
			10,15,20				& 82.0 & 82.0 & 85.8/88.2 & 85.8/88.2 & 83.3/\textbf{86.5} & 83.8/\textbf{86.5} \\ 
			\bottomrule
		\end{tabular}
		}
		\caption{\textbf{Cluster NO. in S3C Loss}. Cluster number is an important impact to affect the performance.}
		\label{tab:result_of_S3C}
	\end{table*}

	Following previous works \cite{Tsai2019Multimodal, Rahman2020Integrating} and the CMU-MultimodalSDK\footnote{https://github.com/A2Zadeh/CMU-MultimodalSDK}, the video feature is extracted via Facet\footnote{iMotions. Facial expression analysis, 2017.} and the acoustic feature is extracted using COVAREP \cite{degottex2014covarep}. The video feature mainly contains 35 facial action units, e.g., facial muscle movement. The acoustic feature mainly includes Mel-frequency cepstral coefficients (MFCCs), pitch tracking and voiced/unvoiced segmenting features, glottal source parameters, peak slope parameters, and maxima dispersion quotients. The video feature dimension $\hat{d}_v$ is 35 for IEMOCAP and CMU-MOSEI, and 47 for CMU-MOSI. The acoustic feature dimension $\hat{d}_a$ is 74 for all three benchmarks. We refer to this version of the feature as \emph{Facet\&COVAREP}.

	For the IEMOCAP, we also compare the video feature extracted by OpenFace\footnote{https://github.com/TadasBaltrusaitis/OpenFace} and the acoustic feature extracted by librosa\footnote{https://github.com/librosa/librosa} to investigate the influence of the unimodality representation. Compared with CMU-MOSI and CMU-MOSEI, each frame of IEMOCAP has two people in the scenario simultaneously, making the judgment difficult. We partition two people according to the layout of the frame and extract the feature separately. The video feature dimension $\hat{d}_v$ is 709 and the acoustic feature dimension $\hat{d}_a$ is 33. We refer to this version of the feature as \emph{OpenFace\&Librosa}.
	\begin{table}[!tp]
		\setlength{\tabcolsep}{8pt}
		\centering
		\scalebox{1.0}{
		\begin{tabular}{ccc}
			\toprule
			\multirow{2}{*}{Modality} & \multicolumn{2}{c}{CMU-MOSI} \\ 
									& BA$\uparrow$ & F1$\uparrow$ \\ \midrule
			T						& 86.4/88.6 & 86.4/88.6 \\ 
			V						& 53.1/54.1 & 52.9/54.0 \\ 
			A						& 54.7/55.0 & 54.1/54.4 \\ 
			T,V						& 86.6/88.9 & 86.5/88.9 \\ 
			T,A						& 87.0/\textbf{89.3} & 87.0/\textbf{89.3} \\  
			V,A						& 54.9/55.4 & 54.9/55.6 \\ 
			T,V,A					& \textbf{87.2/89.3} & \textbf{87.3/89.3} \\ 
			\bottomrule
		\end{tabular}
		}
		\caption{\textbf{Multi-modality Fusion.} Combining different unimodality can improve model performance. T, V, and A mean text, video, and audio modality, respectively. For the BA and F1 of CMU-MOSI and CMU-MOSEI, we report two values: the left side of ``/'' is calculated following \citet{zadeh2018multi}, and the right side is following \citet{Tsai2019Multimodal}.}
		\label{tab:result_of_modality}
	\end{table}
	\begin{table}[tp]
		\setlength{\tabcolsep}{8pt}
		\centering
		\scalebox{1.}{
		\begin{tabular}{lcccccccc}
			\toprule
			\multirow{2}{*}{Features} & \multicolumn{2}{c}{IEMOCAP} \\ 
										& $\overline{\text{Acc}}\uparrow$ & $\overline{\text{F1}}\uparrow$ \\ \midrule
			\emph{Facet\&COVAREP}		& 82.6 & 82.4 \\ 
			\emph{OpenFace\&Librosa}	& \textbf{85.1} & \textbf{85.0} \\
			\bottomrule
		\end{tabular}
		}
		\caption{\textbf{Nonverbal Feature.} Stronger nonverbal features can improve performance.}
		\label{tab:result_of_features}
	\end{table}
	\begin{table*}[tp] 
		\setlength{\tabcolsep}{5pt}
		\centering
		\scalebox{0.80}{ 
		\begin{tabular}{lcccccccccc}
			\toprule
			\multirow{2}{*}{Methods} & \multicolumn{2}{c}{Happy} & \multicolumn{2}{c}{Sad} & \multicolumn{2}{c}{Angry} & \multicolumn{2}{c}{Neutral} & \multicolumn{2}{c}{\textit{Average}} \\ 
										& Acc$\uparrow$ & F1$\uparrow$ & Acc$\uparrow$ & F1$\uparrow$ & Acc$\uparrow$ & F1$\uparrow$ & Acc$\uparrow$ & F1$\uparrow$ & $\overline{\text{Acc}}\uparrow$ & $\overline{\text{F1}}\uparrow$ \\ \midrule
			CTC + EF-LSTM \cite{Tsai2019Multimodal} & 76.2 & 75.7 & 70.2 & 70.5 & 72.7 & 67.1 & 58.1 & 57.4 & 69.3 & 67.7 \\
			LF-LSTM \cite{Tsai2019Multimodal} & 72.5 & 71.8 & 72.9 & 70.4 & 68.6 & 67.9 & 59.6 & 56.2 & 68.4 & 66.6 \\
			CTC + RAVEN \cite{Wang2019Words} & 77.0 & 76.8 & 67.6 & 65.6 & 65.0 & 64.1 & 62.0 & 59.5 & 67.9 & 66.5 \\
			CTC + MCTN \cite{Pham2019Found} & 80.5 & 77.5 & 72.0 & 71.7 & 64.9 & 65.6 & 49.4 & 49.3 & 66.7 & 66.0 \\
			MulT \cite{Tsai2019Multimodal} & 84.8 & 81.9 & 77.7 & 74.1 & 73.9 & 70.2 & 62.5 & 59.7 & 74.7 & 71.5 \\
			PMR \cite{Lv_2021_Progressive} & 86.4 & 83.3 & 78.5 & 75.3 & 75.0 & 71.3 & 63.7 & 60.9 & 75.9 & 72.7 \\
			MTAG \cite{Yang2021MTAG} & - & \textbf{86.0} & - & 79.9 & - & 76.7 & - & 64.1 & - & 76.7 \\
			\midrule
			ScaleVLAD                & \textbf{86.7} & 85.9 & \textbf{84.8} & \textbf{84.6} & 86.8 & 86.9 & \textbf{72.1} & \textbf{72.1} & \textbf{82.6} & \textbf{82.4} \\
			\multicolumn{1}{l}{\quad - w/o multi-scale} 	& 86.6 & 85.7 & 84.1 & 84.2 & 86.7 & 86.9 & 71.5 & 71.3 & 82.2 & 82.0 \\
			\multicolumn{1}{l}{\quad - w/o S3C loss}  & 85.1 & 84.9 & 84.3 & 84.4 & \textbf{88.5} & \textbf{88.3} & 69.4 & 68.5 & 81.8 & 81.5 \\
			\bottomrule
		\end{tabular}
		}
		\caption{\textbf{Sentiment prediction on IEMOCAP (unaligned) dataset.} $\overline{\text{Acc}}$ and $\overline{\text{F1}}$ are the average values. CTC \cite{Graves2006Connectionist} denotes connectionist temporal classification. The results of CTC + EF-LSTM, LF-LSTM, CTC + RAVEN and CTC + MCTN are from \cite{Tsai2019Multimodal}.}
		\label{tab:result_of_IEMOCAP}
	\end{table*}

	\subsection{Experimental Details}
	We initial the text encoder with T5 Base Encoder \cite{Raffel2020t5} in this paper due to its advantages after training with an extensive corpus. We also conduct an ablation study to compare with BERT Base uncased version \cite{Devlin2019BERT}. The rest of the parameters, e.g., Video Transformer, Audio Transformer, and Fusion module, are initialized randomly. The fusion dimension $d_s$ is set to 128. We train the model with the Adam optimizer \cite{kingma2014adam} with a linear schedule. The warmup rate is set to 0.1 based on the total epoch 50. The learning rate is set from \{1e-3, 1e-4, 5e-5, 1e-5\}. The Video Transformer and Audio Transformer are set from \{4, 6\} layers with \{128, 768\} hidden size. The fusion Transformer in Eq. (\ref{fusion_transfomer}) is set with layer 2. The multi-scale parameter $m$ and the number of shared semantic vectors $K$ in Eq. (\ref{eq_wij}) is set from \{1, 2, 3, 10\} and \{8, 10\}, respectively. The cluster $C$ is set from \{10, 15, 20\}. Note these candidate choices are not exact and also can not set with a grid search strategy, so we set them through empirical testing on validation set. The start epoch $s_s3c$ for loss $\mathcal{L}_{s3c}$ is set to 5, the same as the warmup epochs. All hyper-parameters are set according to the performance from the validation set. The batch size is 64 across three datasets. All experiments are carried out on 4 NVIDIA Tesla V100 GPUs.
	\begin{table*}[tp] 
		\setlength{\tabcolsep}{5pt}
		\centering
		\scalebox{0.80}{
		\begin{tabular}{lcccc}
			\toprule
			Methods         & BA$\uparrow$ & F1$\uparrow$ & MAE$\downarrow$ & Corr$\uparrow$ \\ \midrule
			MV-LSTM \cite{Rajagopalan2016Extending} & 73.9/- & 74.0/- & 1.019 &  0.601 \\
			TFN \cite{Zadeh2017Tensor} 		& 73.9/- & 73.4/- & 1.040 & 0.633 \\
			MARN \cite{zadeh2018multi} 		& 77.1/- & 77.0/- & 0.968 & 0.625 \\
			MFN \cite{zadeh2018memory} 		& 77.4/- & 77.3/-  & 0.965 & 0.632 \\
			RMFN \cite{Liang2018Multimodal} & 78.4/- & 78.0/- & 0.922 & 0.681 \\
			RAVEN \cite{Wang2019Words} 		& 78.0/- & -/- & 0.915 & 0.691 \\
			MulT \cite{Tsai2019Multimodal} 	& -/81.1 & -/81.0 & 0.889 & 0.686 \\
			ICCN \cite{sun2020learning}   	& -/83.1 & -/83.0 & 0.862 & 0.714 \\
			PMR \cite{Lv_2021_Progressive} & -/82.4 & -/82.1 & - & - \\
			FMT \cite{Zadeh2019Factorized} 	& 81.5/83.5 & 81.4/83.5 & 0.837 & 0.744 \\
			UniVL \cite{Luo2020UniVL}   	& 83.2/84.6 & 83.3/84.6 & 0.781 & 0.767 \\
			% MISA \cite{Hazarika2020MISA} 
			MISA (Hazarika et al. 2020) 	& 81.8/83.4 & 81.7/83.6 & 0.783 & 0.761 \\
			MAG-BERT \cite{Rahman2020Integrating}	& 84.2/86.1 & 84.1/86.0 & 0.712 & 0.796 \\
			MAG-XLNet \cite{Rahman2020Integrating}	& 85.7/87.9 & 85.6/87.9 & \textbf{0.675} & \textbf{0.821} \\
			Self-MM \cite{Yu2021SelfMM}	& 84.0/86.0 & 84.4/86.0 & 0.713 & 0.798 \\
			MTAG \cite{Yang2021MTAG} & -/82.3 & -/82.1 & 0.866 & 0.722 \\
			\midrule
			ScaleVLAD 	& \textbf{87.2/89.3} & \textbf{87.3/89.3} & 0.684 & 0.819 \\
			\multicolumn{1}{l}{\quad - w/o multi-scale} 	& 86.3/88.6 & 86.2/88.6 & 0.713 & 0.807 \\
			\multicolumn{1}{l}{\quad - w/o S3C loss} 	& 86.0/88.0 & 85.9/88.0 & 0.727 & 0.810 \\
			\midrule
			Human 	& 85.7/- & 87.5/- & 0.710 & 0.820 \\
			\bottomrule
		\end{tabular}
		}
		\caption{\textbf{Sentiment prediction on CMU-MOSI dataset.} For BA and F1, we report two values: the left side of ``/'' is calculated following \citet{zadeh2018multi}, and the right side is following \citet{Tsai2019Multimodal}.}
		\label{tab:result_of_MOSI}
	\end{table*}

	\subsection{Ablation Studies}
	We conduct comprehensive ablation studies on text encoder, key hyper-parameters settings, and features in this section.
	
	\vspace{0.05cm}
	\noindent
	\textbf{Text Encoder.} In Table \ref{tab:result_of_pretrain}, we compare the BERT-Base with the T5-Base. The T5-Base wins on CMU-MOSI and CMU-MOSEI. Besides, it has comparable results on IEMOCAP. Thus, we use T5-Base as our text encoder in our work. We suppose that a larger pretrained model, e.g., T5-Large, can achieve better performance but needs more computational resources. 
	
	\vspace{0.05cm}
	\noindent
	\textbf{Multi-scale Fusion.} In Table \ref{tab:result_of_scale}, we ablate the scale setting of the ScaleVLAD module. The table lists a part of combinations, and we find \{1,2,10\} and \{1,2,3,10\} can achieve better results than others. It proves that fusing different granularities of representation can achieve better performance.
	
	\vspace{0.05cm}
	\noindent
	\textbf{Cluster NO. in S3C Loss.} In Table \ref{tab:result_of_S3C}, we ablate the cluster setting of S3C loss. The table lists a part of combinations, and we find \{10,15\} and \{15,20\} can achieve better results than others. It indicates an appropriate choice of the cluster will keep the feature clustering and thus improve the results.
	
	\vspace{0.05cm}
	\noindent
	\textbf{Multi-modality Fusion.} The results in Table \ref{tab:result_of_modality} prove that multimodal fusion can provide more comprehensive information and capture more emotional characteristics than unimodality.
	
	\vspace{0.05cm}
	\noindent
	\textbf{Nonverbal Feature.} In Table \ref{tab:result_of_features}, different nonverbal features are conducted on IEMOCAP. It shows that more sophisticated features can obtain better results. Further, we suppose that end-to-end training from raw signals instead of the features extracted by off-the-shelf tools can improve more, like video retrieval from \cite{Luo2021CLIP4Clip}.
	\begin{table*}[tp] 
		\setlength{\tabcolsep}{5pt}
		\centering
		\scalebox{0.80}{
		\begin{tabular}{lcccc}
			\toprule
			Methods         & BA$\uparrow$ & F1$\uparrow$ & MAE$\downarrow$ & Corr$\uparrow$ \\ \midrule
			MV-LSTM \cite{Rajagopalan2016Extending} & 76.4/- & 76.4/- & - &  - \\
			MFN \cite{zadeh2018memory} 		& 76.0/- & 76.0/- & - & - \\
			RAVEN \cite{Wang2019Words} 		& 79.1/- & 79.5/- & 0.614 & 0.662 \\
			PMR \cite{Lv_2021_Progressive} & -/83.1 & -/82.8 & - & - \\
			% MISA \cite{Hazarika2020MISA}
			MAG-BERT \cite{Rahman2020Integrating}	& -/84.7 & -/84.5 & - & - \\
			MAG-XLNet \cite{Rahman2020Integrating}	& -/85.6 & -/85.7 & - & - \\
			TFN \cite{Zadeh2017Tensor} 		& -/82.5 & -/82.1 & 0.593 & 0.700 \\
			MulT \cite{Tsai2019Multimodal} 	& -/81.6 & -/81.6 & 0.591 & 0.694 \\
			ICCN \cite{sun2020learning}   	& -/84.2 & -/84.2 & 0.565 & 0.713 \\
			MISA (Hazarika et al. 2020) 	& 83.6/85.5 & 83.8/85.3 & 0.555 & 0.756 \\
			Self-MM \cite{Yu2021SelfMM}	& 82.8/85.2 & 82.5/85.3 & 0.530 & 0.765 \\
			\midrule
			ScaleVLAD 	& \textbf{84.5/86.4} & \textbf{84.7/86.3} & \textbf{0.527} & \textbf{0.781} \\
			\multicolumn{1}{l}{\quad - w/o multi-scale} 	& 83.1/85.8 & 83.3/85.7 & 0.541 & 0.779 \\
			\multicolumn{1}{l}{\quad - w/o S3C loss} 	& 82.7/86.1 & 83.1/86.1 & 0.548 & 0.773 \\
			\bottomrule
		\end{tabular}
		}
		\caption{\textbf{Sentiment prediction on CMU-MOSEI dataset.} For BA and F1, the values on the both sides of ``/'' have the same calculations as Table \ref{tab:result_of_MOSI}.}
		\label{tab:result_of_MOSEI}
	\end{table*}

	\subsection{Comparison to State-of-the-art}
	\label{sec_comparison}
	We compare ScaleVLAD with state-of-the-art methods on IEMOCAP, CMU-MOSI, and CMU-MOSEI, and the results are shown in Table \ref{tab:result_of_IEMOCAP}, Table \ref{tab:result_of_MOSI}, and Table \ref{tab:result_of_MOSEI}, respectively. In summary, 1) the proposed ScaleVLAD outperforms all baselines in all datasets; 2) The ablation on multi-scale fusion and S3C loss proves their effectiveness in all metrics and datasets. Our BERT-based results shown in Table \ref{tab:result_of_pretrain} can also have advantages over the BERT feature-based models, e.g., UniVL \cite{Luo2020UniVL}, MAG-BERT \cite{Rahman2020Integrating}, and Self-MM \cite{Yu2021SelfMM} in Table \ref{tab:result_of_MOSI}. The T5 based feature can improve the performance of IEMOCAP by a significant margin shown in Table \ref{tab:result_of_IEMOCAP}, which proves the strong capability of the pretrained model after training with an extensive corpus in a self-supervised manner.

	\subsection{Qualitative Analysis}
	% Visualizing Analysis on S3C Loss
	\begin{figure*}[hp]
		\centering
		\subfloat[ScaleVLAD w/o S3C loss] {
			\centering
			\includegraphics[width=0.45\textwidth]{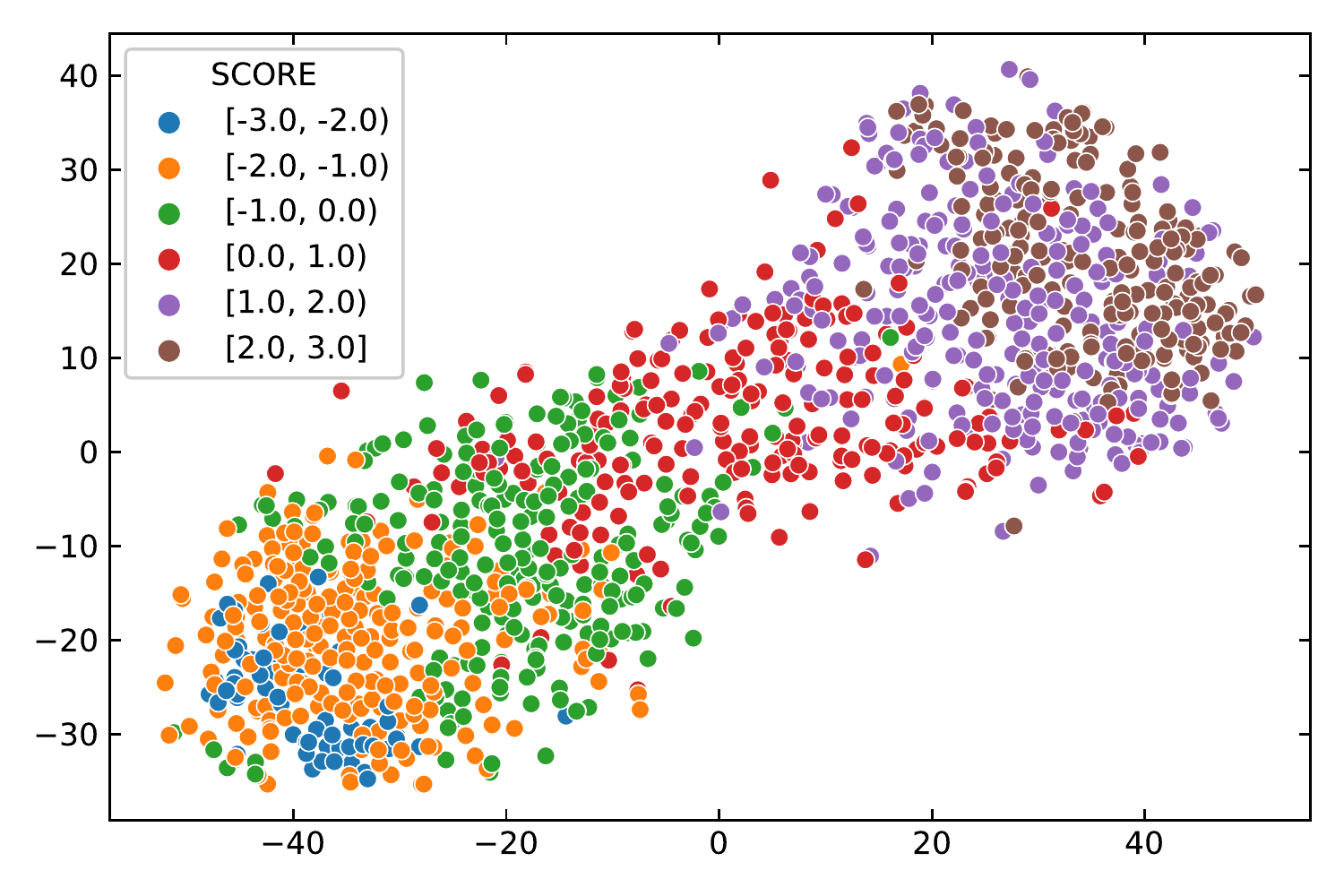} 
			\label{fig:fig_wo_s3c}
		}
		\subfloat[ScaleVLAD w/ S3C loss] {
			\centering
			\includegraphics[width=0.45\textwidth]{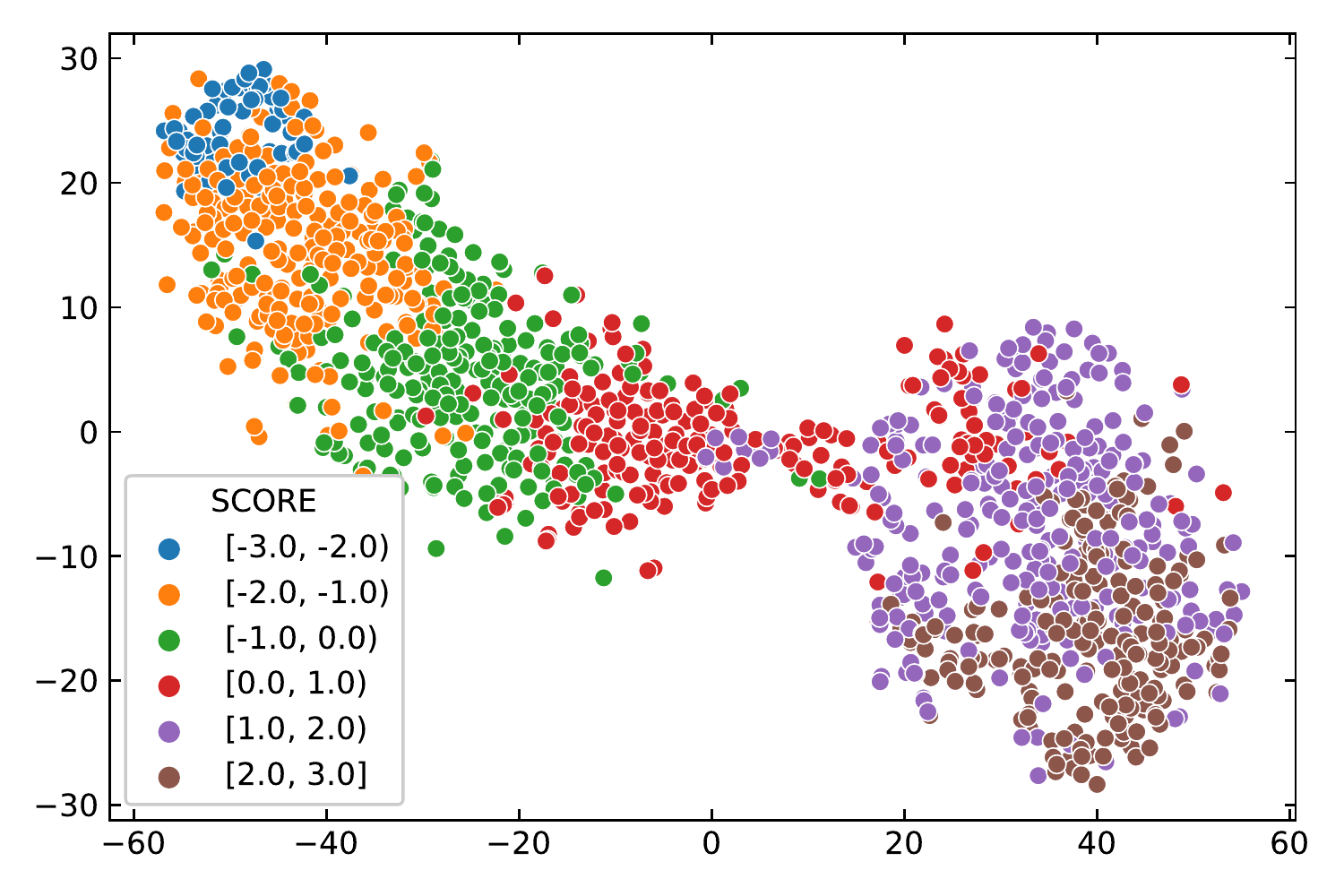} 
			\label{fig:fig_w_s3c}
		}
		\caption{Visualization of the ScaleVLAD w/o and w/ S3C loss in the training set of MOSI using t-SNE projections \cite{vandermaaten08aSNE}.} 
		\label{fig:vis_of_s3c} 
	\end{figure*}
\begin{figure*}[tp]
		\centering
		\includegraphics[width=0.89\textwidth]{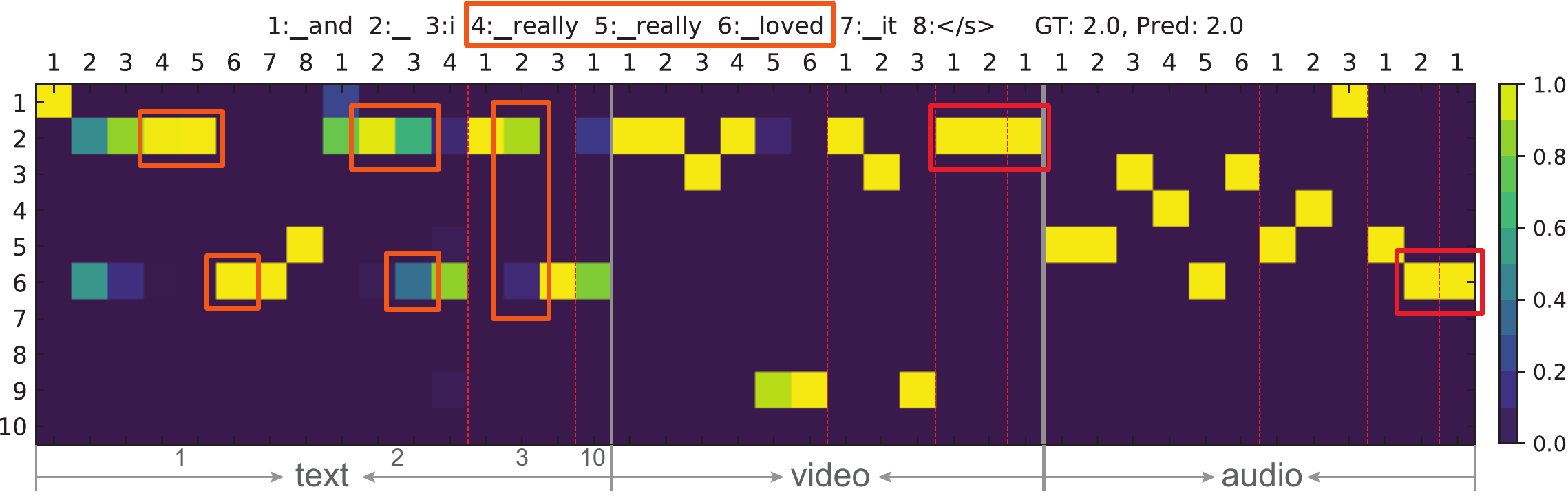}
		\caption{Visualization of the similarity weights from ScaleVLAD module (Eq. (\ref{eq_wij})). The tokens are processed by T5 tokenization. The y-axis means ten shared semantic vectors. The x-axis denotes three blocks: text, video, and audio. Each block has four scales: \{1, 2, 3, 10\}.} 
		\label{fig:vis_of_scalevlad} 
	\end{figure*}

	Figure \ref{fig:vis_of_s3c} displays the visualization of fusion features calculated by Eq. (\ref{output}) on training with S3C loss or not. For a clear observation, we regard the continuous labels as six groups, each having width 1, e.g., $[-3.0, -2.0)$. Figure \ref{fig:fig_w_s3c} illustrates a tight clustering and clearer boundary, e.g., the samples in blue color,  when using S3C loss. It proves the effectiveness of the S3C loss on representation learning. Figure \ref{fig:vis_of_scalevlad} shows the similarity calculated by Eq. (\ref{eq_wij}). The alignment patterns of text, video, and audio with different scales are different and are dynamically learned by the model. In this case, the `really really loved' with the yellow box can be regarded entirely to align with the latent shared semantic vectors. Besides, the video and audio with red boxes, which have bigger scales, i.e., 3 and 10, show consistently shared vectors (NO. 2 and 6) with the text. Through the shared vectors, the model can align and fuse the video and audio representation despite their fuzzy semantic boundaries. We suppose the improvement of the ScaleVLAD is benefits from such a multi-scale alignment.

	\section{Conclusion}
	We proposed a multi-scale fusion method ScaleVLAD and a self-supervised shifted clustering loss to address unaligned multimodal sentiment analysis in this paper. The proposed method considers different granularities of representation through aligning different modalities to trainable latent semantic vectors. Thus, it can remit the fuzzy semantic boundary of unimodality. The proposed self-supervised shifted clustering loss keeps the differentiation of the fusion features via maintaining a momentum updated cluster centers. The extensive experiments on three publicly available datasets demonstrate the effectiveness of the proposed model.
    
    \bibliographystyle{acl_natbib}
	\bibliography{MModalitySA}
\end{document}